\pdfoutput=1

\documentclass[11pt]{article}

\usepackage[preprint]{acl}

\newcommand{\revision}[1]{#1}

\usepackage{xcolor}
\usepackage{soul} 
\usepackage{ulem}
\sethlcolor{lightgray}
\usepackage{balance}

\newcommand{\claim}[1]{%
  ``{\textit{{#1}}}''
}
\newcommand{\name}{FactLens}
\usepackage{makecell}
\usepackage{times}
\usepackage{latexsym}
\usepackage{amsmath}
\usepackage{amsfonts}
\usepackage[T1]{fontenc}

\usepackage[utf8]{inputenc}

\usepackage{microtype}

\usepackage{inconsolata}

\usepackage{graphicx}

\title{\name{}: Benchmarking Fine-Grained Fact Verification}

\author{
  Kushan Mitra\\
  Megagon Labs, USA\\
  \texttt{kushan@megagon.ai}
  \And
  Dan Zhang\\
  Megagon Labs, USA\\
  \texttt{dan\_z@megagon.ai}
  \AND
  Sajjadur Rahman\thanks{Work done while at Megagon Labs.}\\
  Adobe Inc.\\
  \texttt{sajjadurr@adobe.com}
  \And
  Estevam Hruschka\\
  Megagon Labs, USA\\
  \texttt{estevam@megagon.ai}
}

\begin{document}
\maketitle
\begin{abstract}
  Large Language Models (LLMs) have shown impressive capability in language generation and understanding, but their tendency to hallucinate and produce factually incorrect information remains a key limitation. To verify LLM-generated contents and claims from other sources, traditional verification approaches often rely on holistic models that assign a single factuality label to complex claims, potentially obscuring nuanced errors. In this paper, we advocate for a shift towards fine-grained verification, where complex claims are broken down into smaller sub-claims for individual verification, allowing for more precise identification of inaccuracies, improved transparency, and reduced ambiguity in evidence retrieval. However, generating sub-claims poses challenges, such as maintaining context and ensuring semantic equivalence with respect to the original claim.
  We introduce \name{}\footnote{\href{https://github.com/megagonlabs/factlens}{https://github.com/megagonlabs/factlens}}
  , a benchmark for evaluating fine-grained fact verification, with metrics and automated evaluators of sub-claim quality. The benchmark data is manually curated to ensure high-quality ground truth. Our results show alignment between automated \name{} evaluators and human judgments, and we discuss the impact of sub-claim characteristics on the  overall verification performance.
\end{abstract}

\section{Introduction}
Large Language Models (LLMs) have proven to be powerful tools, demonstrating impressive capabilities in language generation and understanding \cite{Touvron2023LLaMAOA, NEURIPS2020_1457c0d6}. However, a well-known limitation of LLMs is their tendency to hallucinate, generating information that is factually incorrect or unsupported by evidence \cite{Ji2022SurveyOH, lin-etal-2022-truthfulqa}. As LLMs become more widespread, especially in applications where factual accuracy is crucial, there has been increasing research on methods to verify the factuality of LLM-generated content as well as claims from other sources.

Previous works on building fact-checking benchmarks focus on generating \texttt{claims} with a ground truth \texttt{label}, and in some cases provide the \texttt{evidence/context} to verify the claim
\cite{Aly2021FEVEROUSFE, schlichtkrull-etal-2024-automated}. Claims are generated using human annotators \cite{Aly2021FEVEROUSFE}, synthetic processes \cite{FatahiBayat2023FLEEKFE, tang-etal-2024-minicheck}, or considering LLM outputs on Question-Answering tasks \cite{wang-etal-2024-factcheck}. To increase the complexity of the fact-checking process, the claims are generated from source data of multiple domains \& modalities, such as Wikipedia text and/or tables \cite{thorne-etal-2018-fever, 2019TabFactA, Aly2021FEVEROUSFE}, Web Pages \cite{schlichtkrull-etal-2024-automated}, Knowledge Graphs \cite{kim-etal-2023-factkg}, online posts/chats \cite{wang-etal-2024-factcheck, li-etal-2024-self}, and QA tasks from various domains such as statistics, finance, legal, etc \cite{jacovi2024coverbench}.


These works also provide baseline fact-checking pipelines, which typically involves two main stages: (1) the retrieval of relevant evidence using Search APIs and multimodal data-lakes \cite{tang2024verifai, schlichtkrull-etal-2024-automated} and (2) the verification of claims based on that evidence using NLI-based, LLM-based and fine-tuned fact-verification models \cite{li-etal-2024-self}. Some works also explore delegating these steps entirely to an LLM-based policy framework \cite{li-etal-2024-self, Peng2023CheckYF}. 

Despite this structured pipeline, most existing methods rely on a holistic verification model, where complex claims are assigned a single factuality label, often obscuring the nuanced nature of the errors or inaccuracies in the claims. \revision{In this work, we echo the sentiments of \citet{wang-etal-2024-factcheck, liu-etal-2020-fine, pan-etal-2023-fact, min-etal-2023-factscore,  si-etal-2024-checkwhy} for a shift towards fine-grained verification of complex claims, where claims are decomposed into smaller, more manageable sub-claims that can be individually verified.} We additionally emphasise on the need to provide evaluation metrics to benchmark such fine-grained verification, and enrich existing benchmarks with fine-grained verification labels.

\begin{figure}[t]
   \includegraphics[width=1.0\linewidth]{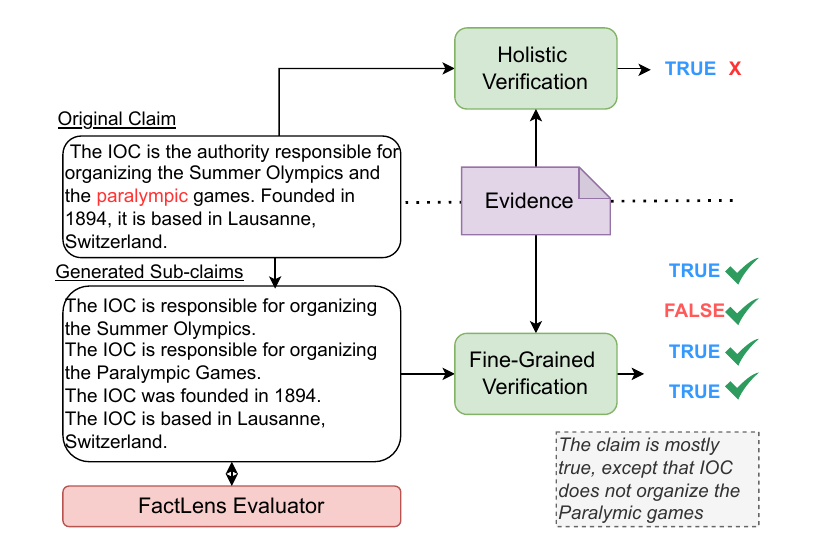} \hfill
  \caption {Examples of holistic fact verification (upper) failed to identify inaccuracies, whereas fine-grained verification (lower) clearly pinpointed the sources of error. In fine-grained verification, the \name{} Evaluator can be used to assess individual sub-claims and identify any alarming signals that may suggest the need for human intervention or regeneration of the sub-claims.}
\label{fig:diagram}
\end{figure}

As shown in Figure \ref{fig:diagram}, the benefits of fine-grained verification are substantial. By breaking down a complex claim into its constituent sub-claims, verification is more precise, allowing for pinpointing exact locations of factual inaccuracies. Additionally, this approach enables more transparent rationalizations and explanations, as each sub-claim can be linked directly to its corresponding evidence or lack thereof. Fine-grained decomposition also narrows the scope of evidence retrieval, making the subsequent verification process more focused and less prone to ambiguity. 

Achieving fine-grained verification, however, presents its own challenges. Decomposing a raw, complex claim into smaller sub-claims is not simply a matter of splitting it into sentences. Poorly constructed sub-claims can introduce a variety of issues: they may lose the context necessary for proper verification, lack atomicity, or misrepresent the original information by either omitting key details or introducing new fabricated ones. Ensuring the quality and verifiability of these sub-claims is, therefore, critical for the overall success of the verification process.

To address these challenges, we introduce \name{}, a benchmark designed specifically for fine-grained fact verification. \name{} provides a novel suite of metrics for evaluating the quality of sub-claim generation and incorporates automated evaluators that combine LLM-based assessments with statistical metrics. The dataset has been manually curated to ensure high-quality sub-claims.

Through empirical evaluation, we demonstrate that our sub-claim evaluators align closely with human judgments. Moreover, our end-to-end evaluation shows that these fine-grained scores correlate strongly with improved downstream verification performance. We also present the results of state-of-the-art models on sub-claim generation, revealing the challenges inherent in this task and the need for further research in this area.





\section{Evaluating Sub-claims with \name{} }
\label{sec:eval}

At the core of fine-grained verification is the decomposition of complex claims into smaller, more specific sub-claims when necessary. The accuracy of the overall verification process depends heavily on the quality of these sub-claims; errors (e.g. oversimplification, omission of important details, or incorrect contextualization) in their formulation can lead to flawed verification outcomes. To detect potential issues early on in the sub-claim generation process, we propose a set of metrics to quantitatively assess sub-claim quality across several dimensions.

\subsection{Evaluation Metrics}
\label{sec:eval}

The decomposed sub-claims should meet criteria below to fully realize benefits of fine-grained verification such as more precise identification of inaccuracies, enhanced transparency, and reduced ambiguity during evidence retrieval.

\begin{table*}[ht]
  \begin{center} 
  \captionsetup{justification=centering}
  \begin{tabular}{lllllllllllll}
  \hline
         Correlation
         & \multicolumn{2}{c}{\textit{atomicity}} & \multicolumn{2}{c}{\textit{sufficiency}} & \multicolumn{2}{c}{\textit{fabrication}} & \multicolumn{2}{c}{\textit{coverage}} & \multicolumn{2}{c}{redundancy}  \\
          & $r$ & $\rho$ & $r$ & $\rho$ & $r$ & $\rho$ & $r$ & $\rho$ & $r$ & $\rho$ \\
  \hline
  \hline
  LLM  &  0.40   &   0.39     &    0.14  &    0.09    & \textbf{0.34}  &   \textbf{0.43}     &  0.43   &    0.45    & \textbf{0.56}    &   \textbf{0.52}      \\
  Statistical   &  \textbf{0.58}   & \textbf{0.59}       & ---    &  ---     &  0.04   &   0.05     & \textbf{0.61}    &   \textbf{0.58}     &   0.10  &      0.12     \\
\hline

  \end{tabular}
  \caption{\label{citation-guide}
      Correlation of \name{} Evaluator scores with Human annotations on synthetic data (Readability is omitted due to its high subjectivity). $r$: Pearson Correlation Score; $\rho$: Spearman Correlation Score. 
    }
  \label{tab:correlation}
  \end{center}
  \end{table*}

\paragraph{Atomicity} Each sub-claim should refer to a single factual unit within the original claim. This ensures if an error occurs, the inaccuracy can be precisely traced back to one or more specific sub-claims. Atomicity measures whether a sub-claim is truly atomic i.e. it focuses on only one relation between a subject and an object. For example, the claim \claim{The International Olympic Committee (IOC) was established on June 23, 1895, in Paris, France} is not atomic, as it makes assertions about both the time and location of the IOC's establishment.

The decomposition process transforms a single claim into a list of sub-claims. It is crucial this transformation is semantically equivalent, ensuring the combined list of sub-claims faithfully represents the original claim and that each can be independently verified. To address this aspect, we propose the metrics \textit{Sufficiency}, \textit{Fabrication}, and \textit{Coverage}.

\paragraph{Sufficiency} To perform fine-grained verification, each sub-claim needs to be independently verifiable. This requires the sub-claims to be properly contextualized to avoid any added ambiguity. \textit{Sufficiency} measures whether the sub-claim is unambiguous and sufficiently contextualized with respect to the original claim. For example, in the original claim \claim{Amanda Bauer attended the University of Cincinnati. The school's nickname is Bearcats.}, the sub-claim \claim{The school's nickname is Bearcats} would be considered low in sufficiency because the reference to the school was omitted in the decomposition, making it ambiguous.

\paragraph{Fabrication} The decomposition process must not introduce additional information or attempt to correct factual errors. This metric is especially important when evaluating LLM decomposers, as LLMs are known to suffer from hallucination or the generation of made-up information. For example, in the original claim \claim{Sydney, the capital of Australia, is known for its Opera House and Harbour Bridge}, a sub-claim \claim{Sydney is the capital of New South Wales, Australia} is considered fabrication. Similarly, with the source claim \claim{Net sales will reach 30 million if the growth rate in 2024 is the same as in 2023}, a sub-claim \claim{The growth rate of 2024 is the same as in 2023} is considered fabrication because it treats a condition as a claim.

\paragraph{Coverage} The list of sub-claims must cover all factual assertions in the original claims, leaving no sub-claims missing. For instance, with the claim \claim{Amanda Bauer attended the University of Cincinnati, whose nickname is Bearcats}, if only one sub-claim is generated as \claim{Amanda Bauer attended the University of Cincinnati}, the coverage will be considered low because the assertion about the university's nickname is missing.

Additionally, some dimensions might not directly affect downstream verifiability and accuracy but capture some nice-to-have characteristics of the sub-claims.

\paragraph{Redundancy} This metric measures whether the sub-claims, as a whole, contain redundant facts. When some sub-claims are semantically repetitive, the distribution of the fact-check units might be skewed. For example, if one erroneous sub-claim is repeated three times, the final judgment could shift from ``mostly correct except for one sub-claim'' to ``more than half of the sub-claims were wrong.'' Furthermore, redundancy also introduces unnecessary costs in terms of time and computing resources.

\paragraph{Readability} This metric assesses how readable the sub-claims are to the end-user and imposes a penalty on unnaturally formed sub-claims.

For each of these metrics, the sub claims are evaluated by assigning a score of `low', `medium' or `high'. For \textit{coverage} and \textit{redundancy}, the scores are assigned at the claim level as we consider the \texttt{sub-claims} as a whole. For all other metrics, scores are assigned to each \texttt{sub-claim} \revision{and then aggregated}. 

\revision{\texttt{Ideal metric values}: For an ideal claim decomposition, we expect the \texttt{sub-claim} to possess high \textit{atomicity}, high \textit{sufficiency}, high \textit{coverage} and high \textit{readability,} while having low \textit{fabrication} and low \textit{redundancy}.}

  

\subsection{\name{} Evaluator}
\label{sec:factlens_evaluator}

\name{} evaluator utilizes an ensemble method of LLM-generated evaluation scores and statistically computed scores (more details in Appendix \ref{appendix: prompt_decomposition} and \ref{appendix: computed_decomposition} respectively). 
We use LLMs as evaluators due to their ability to scale well compared to human evaluators, as well as their reliability and knowledge across diverse domains. However, acknowledging the limitations of LLMs \cite{Bavaresco2024LLMsIO, Stureborg2024LargeLM}, our statistical scores rely on entity and semantic-based computations.

In Table \ref{tab:correlation}, we report the correlation scores of human annotators with the \name{} Evaluator scores, on a synthetic data (more details in Appendix \ref{appendix:agreement} and \ref{appendix:annotators}) that has been carefully curated to cover various types of sub-claim errors. We observe fair to moderate agreement across all dimensions between human evaluations and \name{} Evaluator scores, except for \textit{sufficiency}. The moderate correlation scores can be attributed to the subjectivity involved in judging such metrics. The dependency on contextual information and evidence for assessing the \textit{sufficiency} of a sub-claim contributes to the lower correlation scores for this metric. Nevertheless, our results demonstrate that our computation methods for the \name{} Evaluator align moderately well with human judgments on a dataset with varying sub-claim quality.

\begin{table*}[]
  \centering
  \resizebox{\textwidth}{!}{%
  \begin{tabular}{lllllll}
    \hline
    \textbf{Model} & \textbf{Atomicity} $\uparrow$ & \textbf{Sufficiency} $\uparrow$ & \textbf{Fabrication} $\downarrow$ & \textbf{Coverage} $\uparrow$ & \textbf{Redundancy} $\downarrow$ & \textbf{Readability} $\uparrow$  \\
    \hline
    \textit{Llama-3.1} & 1.87 & 2.85 & 1.01 & 2.88 & 1.09 & 2.96 \\
    \textit{GPT-4o} & 1.82 & 2.85 & 1.02 & 2.89 & 1.15 & 2.95 \\
    \hline
  \end{tabular}
  }
  \caption{
    Measure of sub-claim quality using prompt-based contextualized decomposition. We report average scores on \textit{CoverBench}. Up \& down arrows indicate which metrics should ideally be high ($\sim$ 3) or low ($\sim$ 1) respectively.
  }
  \label{tab:quality}
\end{table*}

\section{\name{}: Benchmarking Fine-grained Verification}

\subsection{Dataset Creation}
\label{sec:dataset_creation}

The \name{} benchmark contains a dataset with ground-truth sub-claims and fine-grained labels. We use 733 instances from \textit{CoverBench}~\cite{jacovi2024coverbench}, a fact-checking benchmark focused on complex claim verification sampled from diverse sources and domains, as the original claims.

We utilize two state-of-the-art LLMs --- \texttt{GPT-4o}~\cite{gpt-4o} and \texttt{LLaMA-3.1}~\cite{llama-31} (details available in Appendix \ref{appendix:decomposistion}) --- to generate candidate sub-claims and measure the quality of these generations using the \name{} Evaluator. To ensure the high quality of the generated sub-claims, we engage human annotators (details provided in the Appendix \ref{appendix:annotators}) to review all sub-claims and manually generate the ground-truth sub-claims, correcting any inaccuracies in the LLM-generated sub-claims (details in Appendix \ref{app:dataset}).

To isolate the benefits of fine-grained verification, we do not perform the step of retrieving evidence or context for each sub-claim. Instead, we use the evidence and context provided in CoverBench, along with the generated sub-claims, to perform fact verification. This approach eliminates variability in the results that could arise from different methods and processes of evidence retrieval.

The next step in fine-grained verification involves using a `verifier' model to fact-check each sub-claim against the provided evidence. In this work, we use \texttt{GPT-4o-mini} as our verifier model across all experiments to maintain consistency.

This verification method enables an enriched fine-grained evaluation. To compare the performance of fine-grained verification labels (for each sub-claim) with the holistic verification label, we aggregate the fine-grained labels as \texttt{false} if at least one of the fine-grained labels is also \texttt{false}; otherwise, the claim is considered \texttt{true}.


\subsection{Claim Decomposition: Model Performance}
\label{appendix:model_perf}

We utilize the prompt defined in Table \ref{tab:prompt1} to decompose claims using \texttt{GPT-4o} and \texttt{Llama-3.1(405B)}. In Table \ref{tab:quality}, we tabulate the evaluation performance of both these models on the claim decomposition task using \name{} Evaluator.

For each instance, we map the `low', `medium', `high' scores (`non-atomic-2', `non-atomic-1', `atomic' for \textit{atomicity}) to numerical values (1, 2, 3 respectively), and report the average for each metric across all 733 instances in the \textit{CoverBench} dataset. 

Both models have similar performance and perform well on the task, as per expected results of having sub-claims which are highly sufficient, low in fabrication, high in coverage, low in redundancy and possess high readability. The \textit{atomicity} scores are far lower, as we qualitatively observe several instances of the type having one subject but multiple objects (`non-atomic-1' which is mapped to a score of 2)

\subsection{Evaluation Results}
\label{sec:eval_results}
\begin{figure}[t]
  \includegraphics[width=1\linewidth]{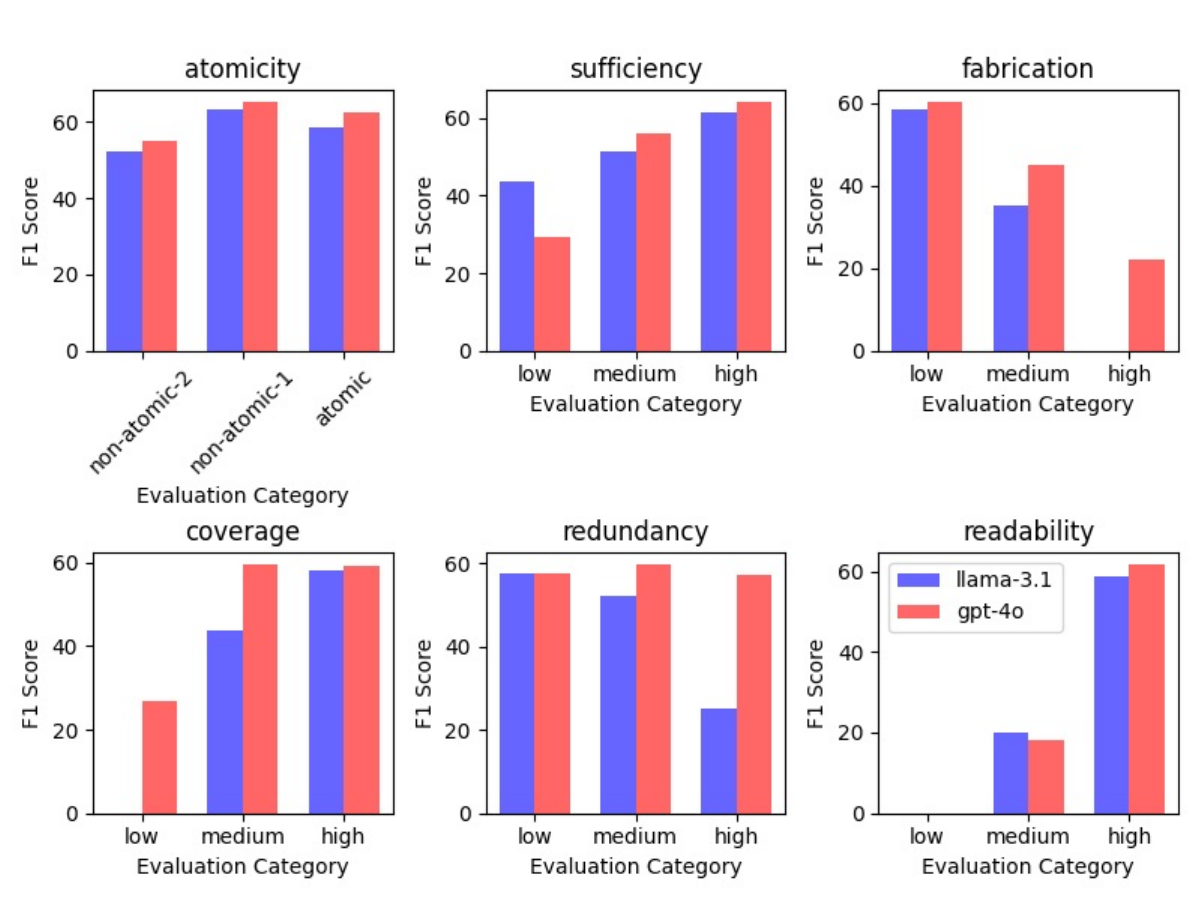} \hfill
  \caption {Impact of sub-claim quality metrics on verification \revision{performance}. See Appendix \ref{appendix: quality} for details.}
\label{fig:metric_quality}
\end{figure}
In fine-grained verification, our \name{} evaluators can act as judges of generated sub-claims, providing early revision signals if the sub-claims might lead to problematic verifications. To illustrate this, we perform an end-to-end evaluation (as shown in Figure \ref{fig:metric_quality}) to highlight how the final verification \revision{performance} is affected by the quality of the sub-claims. For example, we expect high-quality sub-claims to exhibit low \textit{fabrication} scores. We note that for claim decompositions with a \textit{fabrication} score classified as `low,' the \revision{downstream} fact-checking \revision{performance i.e. F1-score,} is higher compared to those with `medium' or `high' fabrication scores.

Similarly, we observe trends where sub-claims with higher \textit{atomicity}, \textit{sufficiency}, \textit{coverage}, and \textit{readability} scores demonstrate better verification performance. Although sub-claims with lower \textit{redundancy} scores perform marginally better, the overall verification performance remains similar. This can be explained by the fact that highly redundant sub-claims may simply repeat claims without negatively impacting the final verification label.

\section{Conclusion}
In this paper, we introduce the benchmark \name{} to evaluate fine-grained claim verification, enriching existing benchmarks. We also identify important metrics for assessing the quality of fine-grained sub-claims and propose an automated evaluator to provide early signals of decomposition failures and evaluate claim decomposition approaches.
\section*{Limitations}


\paragraph{Computation of Metrics} We utilize two methods for computation of \name{} Evaluation metrics: LLM-based and statistically computed. Using LLMs as evaluators/judges is a research field being explored and improved continuously. However, existing works \cite{Stureborg2024LargeLM, Bavaresco2024LLMsIO} have highlighted the limitations of using LLMs in such evaluation tasks, with their scores being skewed and inconsistent. 

To mitigate inconsistency, we provided specific instructions in the prompt (Table \ref{tab:prompt2}) to LLMs. We measured the agreement \& correlation scores of LLMs with human judgement scores, observing fair-moderate agreement across most metrics. Furthermore, we propose our own definitions for computation of the \name{} Evaluator scores.  
However, we acknowledge the limitations in our computational approach as well, with it relying on the method for entity extraction, which may produce variable results. We aim to propose more concrete definitions for these metrics in future works.  

\paragraph{Evidence Retrieval}
To ensure there is no variability in the fact-verification task, in this work we utilize the ground truth evidence which is present in the \textit{CoverBench} dataset. This allows us to solely measure the dependency of fact-verification on the claim-decomposition and sub-claim quality. With our \name{} Benchmark, we provide motivation for fine-grained labels to soon be included across fact-verification benchmarks. In future works, we hope to show how claim decompositions may also improve the evidence search \& retrieval process. 

\paragraph{Fact Verification Models}
Previous works compared different verification models in the fact checking task. \citet{tang-etal-2024-minicheck} contrast the performance of MNLI-based, LLM-based and their proposed fine-tuned models for fact-verification. In this work, we choose to use \texttt{GPT-4o-mini} as our only \texttt{verifier model}, as our aim is not to propose stronger models for verification; but to illustrate the benefits of fine-grained verification even using simpler off-the-shelf \texttt{verifier models}. 

\paragraph{What are Facts?}
Fact extraction is a domain which still has a lot of room for improvement. Some previous works \cite{wang-etal-2024-factcheck} also distinguish between factual claims, opinions and standard sentences. In this work, we utilize the \textit{CoverBench} claims, which itself is obtained from benchmarks which rely on human or synthetic methods for claim generation. It can be noted that different types of factual claims may be generated with such a process, as some claims may center around a claim that is universally true (eg. ``Earth revolves around the Sun."), while other claims are dependent heavily on the context/evidence (eg. ``There are 3 players whose home state is Missouri"). 

Moreover, factual claims in real world-scenarios can often be temporal and domain-dependent in nature. For example, a claim such as \claim{The legal drinking age is 18} is false in a country such as the United States, however may be true in the United Kingdom, indicating domain dependence. Similarly, evidence retrieved for the claim \claim{Barack Obama is the President of the US} is temporal in nature.
We note that our results and experiments are also based on existing benchmarks, which do not account for all real-world scenarios.

\appendix

\section{Claim Decomposition}
\label{appendix:decomposistion}

We utilize few-shot prompting to decompose a claim into sub-claims. Table \ref{tab:prompt1} shows the prompt used to capture the objective of generating sub-claims which are atomic, yet contextualized with enough information preserved from the original claim. 

For the few-shot demonstrations we sample 4 instances from the FEVEROUS dataset \cite{Aly2021FEVEROUSFE}, ensuring no overlap with the \textit{CoverBench} data.
Finally in the prompt, we randomly select 3 of the 4 demonstrations and shuffle the order, to ensure there is no bias. 

\begin{table*}[!htb]
\scriptsize
    \centering
\begin{tabular}{p{5.6in}}
\hline\\
We aim to fact-check a textual claim. To make the fact-checking task simpler, we break down a claim into simpler, atomic sub-claims to fact-check as needed. Note that atomic sub-claims refer to unit claims within the original claim, that refer to a single concept that can be independently verified without having to refer to the original claim. Verification of the sub-claims should not require aggregation of facts or multi-hop reasoning over concepts. However, the sub-claim should have all the contextual information preserved from the original claim. 

\\Your task is to break down a claim into atomic sub-claims for fact checking only if needed. If the original claim itself is a unit claim, do not break it down.

\\For example:
\\\texttt{\{demonstrations\}}

\\Note how each sub claim contains atomic information to fact check and is brief, yet is contextualized with all the information needed from the original claim.

\\Now find the sub claims from the following claim.
\\Claim: \texttt{\{claim\}}
\\Sub\_Claims: \texttt{< your output in form of a list >}
\\
\\$=======================================================================$\\

\\
\\ \hline
\end{tabular}
\caption{{Claim Decomposition Prompt}}
    \label{tab:prompt1}
\end{table*}

\begin{table*}[!htb]
\scriptsize
    \centering
\begin{tabular}{p{5.6in}}
\hline\\
A factual claim can be broken down into atomic, yet contextualized sub-claims which makes it easier to fact check.
You will be provided a claim, and one of the sub-claims which have been extracted from it. Your job is to evaluate this sub-claim on the following metric:

\\\{metric\}

\\Your answer should either be ``low", ``medium" or ``high" based on the metric provided. Please be objective and fair in your evaluation.\\

\\Claim: \{claim\}
\\Sub-Claim: \{sub\_claim\}
\\$=======================================================================$\\

\\\textit{The instructions to calculate each \texttt{metric} is passed one at a time as follows:}

\\``atomicity'': If the sub-claim is atomic i.e. it is simple and centers around only one subject and one object, and the verification does not require aggregation of facts or multihop reasoning over concepts. Label the sub-claim as either ``atomic'' which denotes one subject and one object, or ``non-atomic-1'' which denotes one subject, multiple objects, or ``non-atomic-2'' which denotes multiple subjects

\\``sufficiency'': If the sub-claim itself is sufficient to be fact-checked without the need of any additional contextual information i.e. the sub-claim contains all the required contextual information to be fact-checked independently and is not ambiguous. Your answer should indicate whether the sub-claim has ``low'', ``medium'' or ``high'' sufficiency.

\\"redundancy": If the sub-claims contain redundant or repeated information among them, i.e. multiple semantically equivalent sub-claims. Your answer should indicate whether the sub-claims have ``low'', ``medium'' or ``high'' redundancy.

\\``coverage'': If the set of sub-claims cover all the facts and information made in the original claim. Your answer should indicate whether the sub-claims have ``low'', ``medium'' or ``high'' coverage.

\\``fabrication'': If the sub-claim shows a degree of fabrication  with respect to the original claim i.e. how much new information is added which was not present in the original claim. Note this is not to be judged according to the factuality of the original claims or sub-claims. Your answer should indicate whether the sub-claim has ``low'', ``medium'' or ``high'' fabrication.

\\``readability'': If the sub-claim is readable to an end user. Your answer should indicate whether the sub-claim has ``low'', ``medium'' or ``high'' readability.

\\
\\ \hline
\end{tabular}
\caption{{Prompt for Evaluating Claim Decompositions using LLMs}}
    \label{tab:prompt2}
\end{table*}

We utilize \texttt{GPT-4o} and \texttt{Llama-3.1(405B)} models for this task, with \texttt{temperature} = 0.

\section{Evaluating Claim Decomposition}

To evaluate the sub-claims, our \name{} Evaluator utilizes LLM-generated as well as statistically computed scores. 

\subsection{Prompt for Evaluating Claim Decomposition}
\label{appendix: prompt_decomposition}

In Table \ref{tab:prompt2}, we provide the prompt to LLMs used for evaluating the claim decompositions across the 6 metrics defined in Section \ref{sec:eval}. We provide clear instructions using which LLMs can judge the claim decompositions across different dimensions. For all metrics except ``coverage'' and ``redundancy'', the sub-claims are passed one at a time to obtain sub-claim level evaluation. ``Coverage'' and ``redundancy'' are used to judge the sub-claims as a whole, hence for these metrics, we provide the entire set of sub-claim decompositions for that instance. For ``atomicity" we ask the LLM to output a label as per specific instructions, while for each of the remaining metrics we prompt the LLM to judge the instance with a ``low", ``medium'' or ``high'' score. 

We utilize the smaller and cheaper OpenAI model \texttt{GPT-4o-mini} for this task, with \texttt{temperature} = 0.

\subsection{Statistically Computed Evaluation}
\label{appendix: computed_decomposition}

To compute the \name{} Evaluation metrics using statistical methods we rely on entity-based and semantic-based calculations. Given one instance, with claim $C$, we extract all the (Subject, Object) pairs present within it using \texttt{gpt-4o-mini; temperature = 0}, and from there create a list of $S$ = subjects, and $O$ = objects. After decomposing the claim, we obtain sub-claims $c$ = \{$c_{1}$, $c_{2}$, ..., $c_{n}$\}. 
For all $i$ in [1, n], we extract the subjects $s_{i}$ and object $o_{i}$ lists in a similar manner.
We next follow these definitions to calculate the following metrics:

\paragraph{Atomicity} To measure atomicity, we use an entity-based computation method of comparing the number of subjects and objects involved in the sub-claim $c_{i}$. If $c_{i}$ revolves around only one subject and one object eg. ``\textit{Kurt Cobain was a guitarist}'', it is labeled `atomic'.\\

if $len(s_{i})$ = 1 and $len(o_{i})$ = 1, 
    \\\textit{atomicity} = `atomic'\\

If $c_{i}$ revoles around one subject, but multiple objects eg. ``\textit{Kurt Cobain was a guitarist and a singer}'', it is labeled `non-atomic-1'.\\

    if $len(s_{i})$ = 1 and $len(o_{i})$ \textgreater{} 1, \\\textit{atomicity} = `non-atomic-1'\\

However, if $c_{i}$ revolves around multiple subjects eg. ``\textit{Kurt Cobain was a member of the band Nirvana, which was co-founded with  Krist Novoselic}'', it is labeled `non-atomic-2'.\\    

    if $len(s_{i})$ \textgreater{} 1, \\\textit{atomicity} = `non-atomic-2'\\

\paragraph{Sufficiency}  As sufficiency is a tough metric to judge using semantic techniques, we rely on LLM evaluation scores. 

\paragraph{Fabrication}

To calculate fabrication, we count all subjects in each $s_{i}$ that do not appear in $S$ i.e. subjects in the original claim, and all objects in each $o_i$ that do not appear in $O$ i.e. objects in the original claim.


If the count for $fab$ is equal to 0, i.e. no new entities present in the sub-claims, the \textit{fabrication} is `low'. Based on thresholding values, we assign scores of `medium' or `high' \textit{fabrication}.

\paragraph{Coverage} To measure \textit{coverage}, we check if the entities (subjects and objects) in all sub-claims $c_{i}$ include all the subjects $S$ and objects $O$ present in the original claim.

if $\cup(s_i)$ $\forall$ $i$ = $S$ and $\cup(o_i)$ $\forall$ $i$ = $O$, \\\textit{coverage} = `high'

In case there is no overlap, \textit{coverage} = `low', and for all other cases \textit{coverage} = `medium'

\paragraph{Redundancy}

To calculate redundancy, we use semantic-based technique by measuring BertScore \cite{Zhang2019BERTScoreET} between each pair of sub-claims. If there is high similarity between two 

\[
\textit{red} = \sum_{i=1}^{n} \sum_{j=1}^{n} \mathbb{I}(i \neq j, \textnormal{BertScore}(c_i, c_j) > T)
\]

, where $T$ is a threshold value to find BertScore(.) similarity and $n$ is the number of sub-claims generate for that instance. 

Based on the number of redundant claims found, we assign scores of `low', `medium' and `high'.

\paragraph{Readability} We rely on LLM generated evaluations to measure readability.

\subsection{\name{} Evaluation using Ensemble Method}

As previously mentioned, in Table \ref{tab:correlation} we tabulate the correlation between Human scores and our LLM-generated \& statistically computed scores on the synthetic with varying claim decomposition quality. Based on the results across the metrics, we propose to utilize the statistically computed scores for \textit{atomicity} and \textit{coverage} (as they are better correlated than the LLM-generated scores), while using LLM-generated evaluations for the rest of the metrics in our experiment results in Section \ref{sec:eval_results}.

\subsection{Agreement Scores on Synthetic Data}
\label{appendix:agreement}

\revision{In Table \ref{tab:correlation} we observed fair to moderate correlation between humans and \name{} Evaluator scores through Pearson and Spearman correlation scores. We provide the corresponding \textit{p-values} in Table \ref{tab:p_value}, from which we conclude the correlation scores for \textit{atomicity}, \textit{coverage} (both LLM-Human and Statistical-Human), \textit{fabrication}, and \textit{redundancy} (LLM-Human) are statistically significant.}

\begin{table*}[!htb]
  \begin{center} 
  \captionsetup{justification=centering}
  \resizebox{\textwidth}{!}{%
  \begin{tabular}{lllllllllllll}
  \hline
         P-values
         & \multicolumn{2}{c}{\textit{atomicity}} & \multicolumn{2}{c}{\textit{sufficiency}} & \multicolumn{2}{c}{\textit{fabrication}} & \multicolumn{2}{c}{\textit{coverage}} & \multicolumn{2}{c}{redundancy}  \\
          & $r$ & $\rho$ & $r$ & $\rho$ & $r$ & $\rho$ & $r$ & $\rho$ & $r$ & $\rho$ \\
  \hline
  \hline
  LLM  &  \textbf{4e-10}   &   \textbf{5e-10}     &    0.140  &    0.437    &\textbf{ 5.1e-4 } &   \textbf{2.63e-7}     &  \textbf{2.9e-4 }  &    \textbf{2.4e-4}    & \textbf{3.26e-6}    &   \textbf{7.78e-6}      \\
  Statistical   &  \textbf{1e-9}  & \textbf{1e-9 }      & ---    &  ---     &  0.166  &   0.176    & \textbf{7.7e-8}   &  \textbf{ 1.1e-6}    &   0.404  &      0.332     \\
\hline

  \end{tabular}
  }
  \caption{\label{citation-guide}
      P-values from Table \ref{tab:correlation}: Correlation of \name{} Evaluator scores with Human annotations on synthetic data
    }
  \label{tab:p_value}
  \end{center}
  \end{table*}

In addition to the correlation scores in Table \ref{tab:correlation}, we also report agreement scores between Human annotators and \name Evaluator. We report the \texttt{ordinal} Krippendorff's Alpha score to measure the agreement. We observe fair to moderate agreement across all dimensions except `sufficiency', which can be attributed to the dependency on contextual information and evidence to judge \textit{sufficiency} of a sub-claim. 

The synthetic data is curated using 10 claims from the FEVEROUS benchmark and generating expert-annotated claim decompositions with perturbations. For each claim we generate 7 claim decompositions: one with perfect quality sub-claims, one LLM generated sub-claim, and others using perturbations resulting in lower quality sub-claims corresponding to each of the following 5 metrics: \textit{atomicity}, \textit{sufficiency}, \textit{fabrication}, \textit{coverage} and \textit{redundancy}. We exclude \textit{readability} in the agreement-scores, as it is an extremely subjective metric.

\begin{table}[!htb]
  \centering
  \begin{tabular}{lc}
    \hline
    \textbf{Metric}           & \makecell{\textbf{Krippendorff's}\\ \textbf{Alpha}} \\
    \hline
    Atomicity   &       0.4421 \\
    Sufficiency &      0.0486\\
    Fabrication &       0.4085    \\
    Coverage    &       0.5300     \\
    Redundancy   &      0.4240     \\
    \hline
  \end{tabular}
  \caption{\label{citation-guide}
    Alignment of \name{} Evaluator scores with Human annotations on synthetic data 
  }
\label{tab:agreement}
\end{table}




\section{Expert Annotations}
\label{appendix:annotators}
For human annotations on the synthetic data (Section \ref{sec:factlens_evaluator}) and the creation for the benchmark, we recruited two in-house expert annotators. The annotators are proficient in English, currently based in the United States of America, with at least a graduate-level degree. 
For the task, they were provided the same instructions as the prompt to LLMs in Table \ref{tab:prompt2}. The annotators were clearly explained the objective of the task and how their annotations would be utilized.

The inter-annotator agreement score (Krippendorff Alpha) is high, as tabulated in Table \ref{tab:human_agreement}.

\begin{table}[!htb]
  \centering
  \begin{tabular}{lc}
    \hline
    \textbf{Metric}           & \makecell{\textbf{Inter-Annotator} \\ \textbf{Agreement}}  \\
    \hline
    Atomicity   &       0.73 \\
    Sufficiency &      0.53\\
    Fabrication &       0.54    \\
    Coverage    &       0.86    \\
    Redundancy   &      0.94    \\
    Readability   &      0.96    \\
    \hline
  \end{tabular}
  \caption{\label{citation-guide}
    Inter-Annotator Agreement score
  }
\label{tab:human_agreement}
\end{table}

To measure correlation between the \name{} Evaluator scores and human scores, we do an average of \name{} Evaluator score on a metric with both the annotators, and repeat for all metrics. The human annotators were also used to generate ground-truth sub-claims (Section \ref{sec:dataset_creation}).

\section{Fine-Grained Verification}
\label{appendix:fine-grained-ver}
\begin{figure}[t]
  \includegraphics[width=1.0\linewidth]{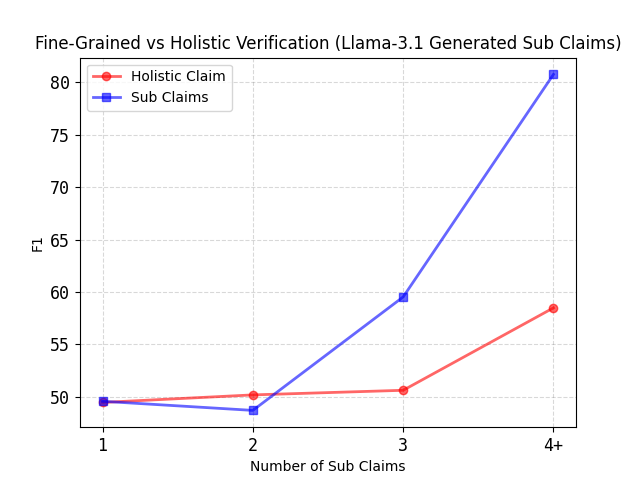} \hfill
  \includegraphics[width=1.0\linewidth]{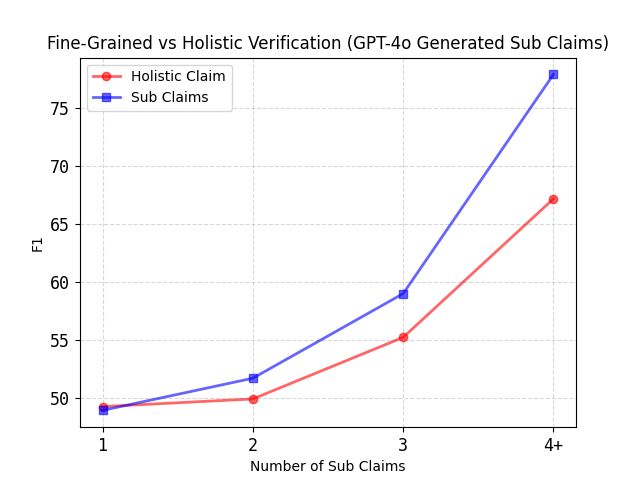}
  \caption {As complexity (i.e. number of sub-claims) increases, the performance of sub-claim decomposition significantly improves. }
\label{fig:sub_claims}
\end{figure}
We study the benefits of fine-grained fact verification compared to verifying the whole claim in Figure \ref{fig:sub_claims}. 
 
In order to perform verification, we utilize \texttt{GPT-4o-mini} to judge if a claim is \texttt{true} or \texttt{false} based on the evidence provided. We obtain the ground truth evidence present in the \textit{CoverBench} dataset. 

In order to show the benefits of fine-grained verification, we compare it with the method of holistically verifying the original claim without decompositions. 

In the first case, we simply pass the original claim $C$ along with the \texttt{evidence} to be verified. In the second case, we pass the claim's decompositions $c$ = \{$c_{1}$, $c_{2}$, ..., $c_{n}$\} one at a time. For each $c_{i}$ we obtain a verification label, and then aggregate the labels for that instance. If any one sub-claim is judged \texttt{false} the whole instance is marked \texttt{false}, otherwise \texttt{true}. 

We contrast the performance of the fine-grained verification with holistic verification in Figure \ref{fig:sub_claims}. Here, we assume the number of sub-claims of an instance is indicative of how complex the claim is.

We observe that as the complexity (number of sub-claims) increase, the performance of the fine-grained verification method significantly increases compared to holistic verification.

\section{Impact of Sub Claim Quality on Verification}
\label{appendix: quality}
\begin{figure}[!htb]
  \includegraphics[width=1.0\linewidth]{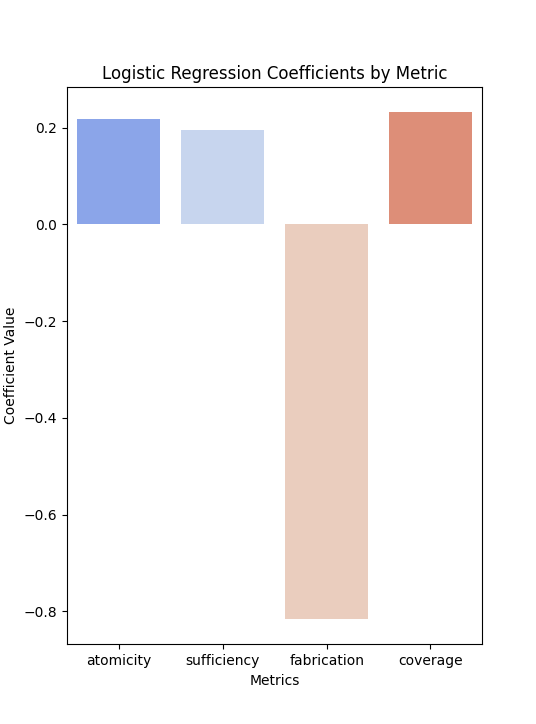} \hfill
  \caption {Fine-Grained Verification: Logistic Regression Coefficients for Metrics}
\label{fig:lr}
\end{figure}

In Figure \ref{fig:metric_quality}, we showed how sub-claim quality impacts the end-to-end verification result. To truly understand the benefits of fine-grained decompositions and the \name{} metrics, we only consider those instances for which the number of sub-claims was greater than 1. Here, we illustrate further using qualitative examples and weights of a logistic regression model to show the influence of \name{} Evaluator Metrics on fine-grained verification. 

To deeper understand how each metric influences the final verification, we fit a logistic regression model on the \name{} Evaluator scores on \textit{CoverBench}. We specifically study the impact of the metrics \textit{atomicity}, \textit{sufficiency}, \textit{fabrication} and \textit{coverage} as we expect them to influence the final verification more than the ``nice-to-have" metrics: \textit{redundancy} and \textit{readability}. 


We also conducted an analysis to understand how the scores can collectively predict the final verification accuracy by fitting a logistic regression model and examining the coefficients associated with each metric. Combining the four metrics—atomicity, fabrication, coverage, and sufficiency—we achieved a prediction F1 score of 0.71, despite potential noise in the retrieval and verification steps.

\begin{figure}[ht]
  \includegraphics[width=1.0\linewidth]{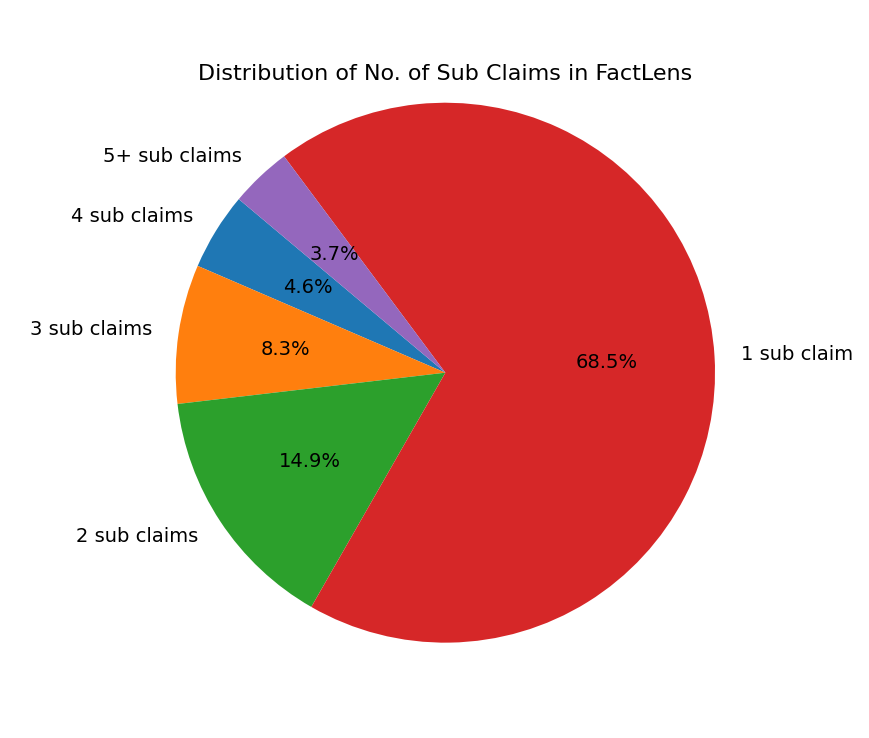} \hfill
  \caption {Distribution of Number of Sub Claims in \name{} Benchmark}
\label{fig:pie}
\end{figure}

From Figure \ref{fig:lr}, we observe that \textit{fabrication} has the highest weight in magnitude, implying most influence in predicting whether the final label matches with ground truth. We see a negative weight for \textit{fabrication} which is expected as lower \textit{fabrication} indicates better quality sub-claims which in turn should have a positive effect on verification. \textit{Atomicity}, \textit{sufficiency} and \textit{coverage} show a positive weight, as highly atomic, highly sufficient and high coverage sub-claims are expected to influence verification positively. 

\begin{table*}[!htb]
\scriptsize
    \centering
\begin{tabular}{p{6in}}
\hline\\
\textbf{Claim}: Fresh water crustaceans Aeglidae are classified as Malocostraca and Decapoda.

\\\textbf{Gold Label}: \emph{False}

\\\textbf{Evidence}: [ground truth evidence from \textit{CoverBench}\ldots]

\\\textbf{Sub Claims}: [`Fresh water crustaceans Aeglidae are classified as Malocostraca', `Fresh water crustaceans Aeglidae are classified as Decapoda']

\\\textbf{\name{} Evaluation}:
\begin{itemize}
    \item \textit{Atomicity}: [`atomic', `atomic']
    \item \textit{Sufficiency}: [`high', `high']
    \item \textit{Fabrication}: [`low', `low']
    \item \textit{Coverage}: `high'
    \item \textit{Redundancy}: `low'
    \item \textit{Readability}: `high'
\end{itemize}

\\\textbf{Fine-Grained Verification Labels}: [True, False]
\\\textbf{Aggregated Fine-Grained Verification Label}: \emph{False}
\\
\\$=======================================================================$\\
\\\textbf{Claim}: In addition to co-starring in a Ken Ludwig musical, Jeffry Denman has worked with notables such as Mel Brooks,  and has been called "a natural scene stealer" by The Houston Chronicle.

\\\textbf{Gold Label}: \emph{False} 

\\\textbf{Evidence}: [ground truth evidence from \textit{CoverBench}\ldots]

\\\textbf{Sub-Claims}: [`Jeffry Denman co-starred in a Ken Ludwig musical', `Jeffry Denman has worked with Mel Brooks', `Jeffry Denman has been called a natural scene stealer by The Houston Chronicle']

\\\textbf{\name{} Evaluation}:
\begin{itemize}
    \item \textit{Atomicity}: [`atomic', `atomic', `non-atomic-2']
    \item \textit{Sufficiency}: [`high', `high', `high']
    \item \textit{Fabrication}: [`low', `low', `low']
    \item \textit{Coverage}: `medium'
    \item \textit{Redundancy}: `low'
    \item \textit{Readability}: [`high', `high', `high']
\end{itemize}

\\\textbf{Fine-Grained Verification Labels}: [True, True, True] 
\\\textbf{Aggregated Fine-Grained Verification Label}: \emph{True}
\\
\\ \hline
\end{tabular}
\caption{{Examples of how sub claim quality impacts verification performance}}
    \label{tab:qualitative}
\end{table*}

We also highlight some qualitative examples in Table \ref{tab:qualitative} to show how sub-claim quality impacts fine-grained verification. In the first instance, with perfect sub-claim quality, the fine-grained verification correctly predicts the ground truth label. In the second instance, we see \textit{coverage} as `medium' and imperfect \textit{atomicity} score whereby the verifier eventually predicts an incorrect label.
\balance

\section{FactLens Dataset Characteristics}
\label{app:dataset}
Our FactLens benchmarks consists of 733 instances from \textit{CoverBench} with ground truth decompositions curated using LLMs and humans, and fine-grained labels as mentioned in Section \ref{sec:dataset_creation}. In Figure \ref{fig:pie} we note the distribution of the number of sub-claims in the dataset.

\end{document}